\documentclass{article}

\PassOptionsToPackage{numbers, compress}{natbib}

\usepackage[preprint]{neurips_2024}

\usepackage[utf8]{inputenc} 
\usepackage[T1]{fontenc}    
\usepackage{hyperref}       
\usepackage{url}            
\usepackage{booktabs}       
\usepackage{amsfonts}       
\usepackage{nicefrac}       
\usepackage{microtype}      
\usepackage{xcolor}         
\usepackage{multirow}
\usepackage{graphicx}
\usepackage{CJKutf8}

\title{CollectiveSFT: Scaling Large Language Models for Chinese Medical Benchmark with Collective Instructions in Healthcare}

\author{%
  Jingwei Zhu \\
  School of Software Engineering \\
  University of Science and Technology of China\\
  Hefei, China \\
  \texttt{jingweizhu@mail.ustc.edu.cn} \\
   \And
   Minghuan Tan \\
   Shenzhen Institute of Advanced Technology\\
   Chinese Academy of Sciences  \\
   Shenzhen, China \\
   \texttt{mh.tan@siat.ac.cn} \\
   \AND
   Min Yang \\
   Shenzhen Institute of Advanced Technology\\
   Chinese Academy of Sciences \\
   Shenzhen, China \\
   \texttt{min.yang@siat.ac.cn} \\
   \And
   Ruixue Li \\
   Xiangshui County Party School \\
   Yancheng, China \\
   \texttt{career\_liruixue@163.com} \\
   \And
   Hamid Alinejad-Rokny \\
 UNSW BioMedical Machine Learning Lab (BML)\\
 School of Biomedical Engineering\\
 UNSW, Sydney, Australia\\
   \texttt{h.alinejad@unsw.edu.au} \\
}

\begin{document}

\maketitle

\begin{abstract}
The rapid progress in Large Language Models (LLMs) has prompted the creation of numerous benchmarks to evaluate their capabilities. This study focuses on the Comprehensive Medical Benchmark in Chinese (CMB)~\cite{wang2023cmb}, showcasing how dataset diversity and distribution in supervised fine-tuning (SFT) may enhance LLM performance. Remarkably, We successfully trained a smaller base model to achieve scores comparable to larger models, indicating that a diverse and well-distributed dataset can optimize performance regardless of model size. This study suggests that even smaller models may reach high performance levels with carefully curated and varied datasets. By integrating a wide range of instructional content, our approach addresses potential issues such as data quality inconsistencies. Our results imply that a broader spectrum of training data may enhance a model’s ability to generalize and perform effectively across different medical scenarios, highlighting the importance of dataset quality and diversity in fine-tuning processes.\footnote{\url{https://github.com/CAS-SIAT-XinHai/CollectiveSFT}}
\end{abstract}

\begin{CJK*}{UTF8}{gbsn}
\section{Introduction}
With the rapid development of Large Language Models (LLMs), there is increasing interest in applying LLMs to the physical health domain. Due to the specialized nature of physical health, LLMs need to acquire extensive medical knowledge, ensure accuracy, and exhibit patience when interacting with patients. To evaluate the knowledge and accuracy of LLMs in this domain, various medical benchmarks have been established. Some models have achieved impressive scores, demonstrating their potential as basic doctor assistants for daily use.

Despite these advancements, several major concerns remain regarding the instructions used for fine-tuning these models. Firstly, the diversity and distribution of instructions may still be limited. As highlighted by Zheng et al.~\cite{zheng2023building}, the effectiveness of fine-tuning is heavily influenced by the variety and richness of the instruction sets used.

To address this issue, we propose integrating a diverse array of instruction types and related domains into our fine-tuning dataset. Our approach involves collecting instructions from multiple question types and ensuring a comprehensive representation of different domains. Specifically, we focus on creating a dataset that includes real-world dialogue reconstructions, consultation records from medical forums, and various other sources. This comprehensive approach aims to enhance the model’s performance across different medical scenarios.

In this work, we explore the potential of supervised fine-tuning (SFT) in improving the performance of a smaller model in the medical domain. By utilizing a diverse and well-distributed dataset, we aim to demonstrate that even a smaller model can achieve competitive performance in specialized tasks. Our experiments highlight the importance of dataset quality in fine-tuning processes and show that a well-curated dataset can significantly enhance a model’s capabilities, even with limited parameters.


\section{Related Work}

\subsection{Instruction Tuning}

Instruction tuning is a highly effective approach for improving the performance of language models on unseen tasks in zero-shot or few-shot scenarios~\cite{wei2022finetuned}. This method involves training models with a variety of instructions, enabling them to better understand and execute tasks they have not been explicitly trained on. 

Natural Instructions~\cite{mishra-etal-2022-cross} represents an effort to create a comprehensive set of human-crafted instructions designed to enhance model performance across a wide range of tasks. These instructions serve as a valuable resource for fine-tuning models to perform well in diverse applications. Building on this concept, Super-NaturalInstructions~\cite{wang-etal-2022-super} expands the scope by including even more detailed and varied instructions, further improving the robustness and adaptability of language models.

To address the issue of limited diversity in human-crafted instructions, Unnatural Instructions~\cite{honovich-etal-2023-unnatural} introduces a vast dataset of imaginative and varied instructions collected with minimal human effort. This innovative approach leverages automated methods to generate a rich and diverse set of instructions, significantly enhancing the model's ability to handle a wider array of tasks with improved accuracy and efficiency.

\subsection{Open-Source Medical Models}

In the realm of medical LLMs, several notable open-source projects have emerged, such as HuatuoGPT~\cite{huatuogpt-2023} and BenTsao~\cite{wang2023huatuo}. These models are designed to assist in medical consultations and diagnostics by leveraging large-scale medical dialogues and literature.

HuatuoGPT and BenTsao~\cite{du2023calla} have undertaken the task of collecting extensive medical dialogue datasets. They use advanced language models like GPT-4 to reconstruct these dialogues into question-answer pairs for model training. This method aims to improve the models' understanding of medical consultations and enhance their ability to provide accurate and relevant responses.

However, these models also come with notable limitations. One major concern is the risk of overfitting to specific datasets, which can limit their generalizability to new, unseen medical scenarios. The reliance on reconstructed dialogues might lead to inconsistencies in data quality, affecting the robustness of the models' responses.

These challenges highlight the need for ongoing refinement and evaluation of open-source medical models. A key area of focus should be the diversity and distribution of datasets used during fine-tuning. Ensuring a wide variety of instructions and data sources may enhance the model's ability to generalize and perform effectively across various medical tasks. By carefully curating and diversifying the datasets, it is possible to develop more robust and versatile medical LLMs, capable of providing reliable and comprehensive support in healthcare settings. Our work aims to address these issues, striving to improve the overall performance of medical LLMs through strategic dataset diversification.

\section{Collective Instruction Set}

\subsection{Data Collection}

The datasets we gather encompass various types, from conversations to question-answering pairs. While we primarily focus on English and Chinese datasets, we also acknowledge the availability of healthcare datasets in other languages, such as HeadQA~\cite{vilares-gomez-rodriguez-2019-head} in Spanish and FrenchMedMCQA~\cite{labrak-etal-2022-frenchmedmcqa} in French. 

Our review of publicly accessible datasets indicated that many formats are unsuitable for model fine-tuning due to inconsistencies in structure, detail levels, and annotation standards. To tackle these issues, we decided to standardize all datasets into the Alpaca format~\cite{alpaca}. This format includes fields for instruction, input, and output, as well as optional fields for system prompts and history, tailored for specific use cases. By adopting a standardized format, we ensure consistent data processing, enhancing its effectiveness for training and fine-tuning models.

Reconstructing the datasets involves several steps. First, we extract relevant information from each dataset, preserving key details. Then, we reformat this information into the Alpaca structure, which entails defining clear instructions for the model, specifying inputs, and providing expected outputs. For conversational data, we include history fields to maintain context across dialogue turns.

Table~\ref{tab:open-data} summarizes all collected data, detailing their language, style, topic size, and instruction size.By aligning diverse datasets into a single, coherent format, we facilitate more effective training processes and enhance the models' ability to generalize across different medical tasks.

In addition to reformatting existing datasets, we also aim to expand our collection with new data sources. This involves curating data from medical forums, academic publications, and other relevant repositories. This ongoing effort ensures our models remain relevant and effective in real-world medical applications.

Moreover, incorporating diverse datasets helps mitigate biases present in individual data sources. By integrating data from various origins and languages, we create a more balanced and comprehensive training environment. This diversity is essential for developing robust, reliable models capable of providing accurate medical advice across different contexts and populations.

\begin{table*}[!htp]\centering
\small
\setlength{\tabcolsep}{5pt} 
\renewcommand{\arraystretch}{1.5} 
\begin{tabular*}{\textwidth}{@{\extracolsep{\fill}} llcrc @{}}
\toprule
Language & Dataset Name & Style & Topic Size & Instruction Size \\\midrule
\multirow{3}{*}{English}
& PubMedQA~\cite{jin-etal-2019-pubmedqa} & QA & 273,518 & 273,518 \\
& MedMCQA~\cite{pmlr-v174-pal22a} & MCQA & 182,822 & 182,822 \\
& HeadQA~\cite{vilares-gomez-rodriguez-2019-head} & QA & 2,657 & 2,657 \\
\cmidrule{2-5}
& Total & & 458,997 & 458,997 \\\cmidrule{1-5}
\multirow{9}{*}{Chinese}
& cMedQA2~\cite{8548603} & QA & 100,000 & 188,783 \\
& cMedDialogue~\cite{cMedDialogue} & QA & 792,099 & 792,099 \\
& webMedQA~\cite{he2019applying} & QA & 252,850 & 50,570 \\
& MedicalDialog~\cite{he2020meddialog} & Dialogue & 2,725,989 & 4,503,475 \\
& CMID~\cite{Chen2020cmid} & NER & 12,254 & 11,786 \\
& NLPEC~\cite{li-etal-2020-towards} & MCQA & 18,703 & 18,703 \\
& CMB~\cite{wang2023cmb} & MCQA & 269,359 & 269,359 \\
& MLEC-QA~\cite{li-etal-2021-mlec} & MCQA & 108,988 & 108,988 \\
& DISCMed~\cite{bao2023discmedllm} & Dialogue & 464,898 & 1,362,307 \\
\cmidrule{2-5}
& Total & & 4,745,140 & 7,306,070 \\\bottomrule
\end{tabular*}
\caption{Public medical datasets used for fine-tuning our model. The table shows their size with original format and number of instructions constructed for this work.}
\label{tab:open-data}
\end{table*}

\subsection{Instruction Set Construction}

We construct instructions based on the data types of the collected datasets, ensuring that each type is processed into a unified format that the language models can effectively utilize. This standardization is crucial for maintaining consistency and clarity across different data sources, which is essential for optimizing the model's performance. The following sections detail the strategies used to process various formats of datasets into a standardized format.

\paragraph{\textbf{Multiple-Choice Question Answering}}
For the MCQA format, we use a consistent method to process the data. The instruction field typically contains background information and descriptions about the source of the question, which helps the LLM understand the context better. The input field combines the original question with all the answer options. The output field provides the correct answer, along with an explanation if available in the dataset.

\paragraph{\textbf{Question Answering}}
The QA format is simpler compared to other formats. We leave the input field blank and fill the instruction field with the original question and the output field with the corresponding answer.

\paragraph{\textbf{Dialogue}}
The dialogue format differs slightly from others due to the nature of conversational data. In this case, we include an additional field named "history" that contains the entire chat history up to that point. The instruction field contains the current question, the input field is left blank, and the output field provides the response. This approach helps the LLM understand the context of the ongoing conversation.

\paragraph{\textbf{Sequence Labeling}}
For sequence labeling, specifically in Named Entity Recognition (NER) tasks, we set the instruction field to request an analysis of specific noun entities and the intent of the description. The input field contains the original content, while the output field consolidates all identified noun entities into a new description that captures the intended meaning. This method aids the LLM in recognizing and understanding specialized terminology in the medical domain.

By standardizing these diverse data formats into a single instructional framework, we ensure consistency and clarity in training. This approach enhances the LLM's ability to generalize and perform effectively across various medical tasks, leading to more reliable and robust models.

\section{Experiments}

\subsection{Hyperparameter Optimization}
We employ advanced tools like LLaMA-Factory~\cite{zheng2024llamafactory} to fine-tune our models, exploring various hyperparameters such as cut-off length, epoch count, and learning rate. These parameters are crucial for the models' performance and efficiency.

For our fine-tuning base model, we have selected the InternLM2.5-7B base model~\cite{cai2024internlm2} due to its outstanding reasoning capabilities. This model stands out for its ability to handle complex tasks with high accuracy and efficiency. Additionally, the 7B parameter size is particularly advantageous as it strikes a balance between performance and resource requirements. This size is common for personal deployment because it does not demand extensive computational resources, making it accessible for a wider range of applications, including those with limited hardware. By choosing the InternLM2.5-7B base model, we aim to leverage its strengths in reasoning while maintaining feasibility for personal and small-scale deployments, ensuring that our fine-tuning processes are both effective and practical.

Our experiments indicate that cut-off length profoundly affects the model's performance. Specifically, a shorter cut-off length yields better results with the same dataset. This improvement is due to the dataset's average length; shorter cut-off lengths help the model capture essential information within each instance, enhancing output accuracy and relevance.

In benchmark scenarios, particularly with multiple-choice questions, a slightly shorter cut-off length proves beneficial. For instance, CMB Exam emphasizes accuracy in answering specific questions over conversational abilities. By aligning the cut-off length with the dataset's average length, we boost the model's efficiency and accuracy for these specialized tasks. Shorter cut-off lengths enable the model to concentrate on the core content of questions and options, improving its ability to select correct answers. Adjusting other hyperparameters like epoch count and learning rate in tandem with cut-off length further refines performance. A higher epoch count allows the model to learn more comprehensively from the training data, while a well-tuned learning rate ensures optimal convergence without overshooting or getting trapped in local minima.

Overall, our hyperparameter optimization strategy balances these parameters to achieve peak performance for specific applications. Through systematic experimentation with different settings, we fine-tune our models to excel in their tasks, ensuring reliable and effective performance in real-world medical applications.

\subsection{Performance over CMB Benchmark}
We achieve an outstanding score in the CMB using a remarkably small model as shown in Table ~\ref{tab:model-performance}, significantly smaller than any other model at the top of the benchmark. This achievement can be attributed to the diversity and distribution of our dataset. Our results demonstrate that the quality of the dataset is the most critical factor influencing the performance of model fine-tuning.

By using a wide variety of data formats and sources, we create a training set that is rich and representative of diverse medical scenarios. This strategy allows our smaller model to generalize better and perform effectively across different tasks within the CMB. The success of our fine-tuning process shows the importance of dataset diversity and demonstrates that even with fewer model parameters, top performance can be achieved through careful dataset selection and distribution.

Furthermore, our findings challenge the conventional belief that larger models are inherently superior. Instead, they emphasize that a well-curated and diverse dataset can significantly enhance a model’s capabilities, enabling smaller models to compete with and even surpass larger ones. This has important implications for the development of efficient, resource-conserving models that do not compromise on performance.

\begin{table*}[!htp]
\centering
\scriptsize 

\resizebox{\textwidth}{!}{
\begin{tabular*}{\textwidth}{@{\extracolsep{\fill}} lccccccc @{}}
\toprule
Model & Total Avg. & \begin{tabular}[c]{@{}c@{}}Training\\ Grad.\end{tabular} & \begin{tabular}[c]{@{}c@{}}Nursing\\ Exam\end{tabular} & \begin{tabular}[c]{@{}c@{}}Pharm.\\ Exam\end{tabular} & \begin{tabular}[c]{@{}c@{}}Med. Tech.\\ Exam\end{tabular} & \begin{tabular}[c]{@{}c@{}}Prof.\\ Knowledge\end{tabular} & \begin{tabular}[c]{@{}c@{}}Med.\\ Postgrad.\end{tabular} \\
\midrule
\textbf{\underline{CollectiveSFT-7B}} & \textbf{\underline{77.05}} & 83.00 & \textbf{\underline{85.75}} & \textbf{\underline{79.25}} & \textbf{\underline{72.50}} & 90.25 & \textbf{\underline{80.25}} \\
InternLM2.5-7B~\cite{cai2024internlm2} & 71.40 & 75.80 & 78.13 & 68.28 & 70.92 & 65.00 & 72.19 \\
HuatuoGPTII-34B~\cite{chen2023huatuogptiionestagetrainingmedical} & 76.80 & 82.50 & 75.50 & 73.25 & 68.75 & 87.75 & 77.00 \\
Qwen-72B-Chat~\cite{qwen} & 74.38 & \textbf{\underline{88.00}} & 75.00 & 77.00 & 70.25 & \textbf{\underline{94.25}} & 65.50 \\
Yi-34B-Chat~\cite{ai2024yi} & 69.17 & 78.75 & 69.50 & 69.75 & 63.75 & 87.00 & 56.50 \\
AntGLM-Med-10~\cite{li2024beginnerexpertmodelingmedical} & 64.09 & 81.75 & 62.00 & 63.75 & 60.25 & 82.50 & 64.50 \\
GPT-4~\cite{openai2023gpt4} & 59.46 & 64.50 & 60.75 & 39.50 & 57.00 & 77.50 & 61.25 \\
HuatuoGPTII-7B~\cite{chen2023huatuogptiionestagetrainingmedical} & 59.00 & 70.75 & 64.75 & 60.00 & 57.75 & 70.25 & 53.75 \\
Qwen-14B-Chat~\cite{qwen} & 57.64 & 69.00 & 60.50 & 51.25 & 51.75 & 73.00 & 50.00 \\
Baichuan2-13B-Chat~\cite{baichuan2023baichuan2} & 48.87 & 56.50 & 47.75 & 44.50 & 45.50 & 63.25 & 39.25 \\
Qwen-7B-Chat~\cite{qwen} & 46.58 & 56.25 & 46.00 & 42.00 & 37.25 & 63.50 & 39.50 \\
ChatGLM2-6B~\cite{du2022glmgenerallanguagemodel} & 45.05 & 48.25 & 47.25 & 43.75 & 43.00 & 54.25 & 42.25 \\
\bottomrule
\end{tabular*}
}
\caption{Performance Comparison of Some Open-source Medical Models focusing on specific exam scores and overall averages.\protect\footnotemark[1]}
\label{tab:model-performance}
\end{table*}

\footnotetext[1]{Only open-source models were selected, excluding closed-source models. Data retrieved from CMB leaderboard on July 24, 2024. (\url{https://cmedbenchmark.llmzoo.com/static/leaderboard.html})}

\section{Discussion and Conclusion}

In this article, we have highlighted the potential of using diverse datasets to improve model performance using SFT. Our findings suggest that incorporating a variety of data types is an effective way to enhance the capabilities of models, achieving better performance with fewer GPU resources.

Our study also uncovered some limitations associated with this method. One notable issue is that while the fine-tuned smaller models excel at answering multiple-choice questions accurately and effectively, they may lose some of their conversational abilities. This loss means that although the models perform well on specific tasks like MCQA, they struggle to maintain engaging and coherent conversations with users during interactive sessions. This trade-off between specialized task performance and general conversational ability is an important consideration for the application in real-world scenarios.

Additionally, we observed common problems associated with smaller models, such as hallucination. Hallucination refers to the generation of plausible but incorrect or nonsensical information by the model. This issue can undermine the reliability of the model's responses and poses a significant challenge for its deployment in sensitive domains like healthcare, where accuracy is paramount.

In conclusion, while the use of diverse datasets in supervised fine-tuning offers a promising pathway for quickly enhancing a model's knowledge base and task-specific performance, it also presents several challenges that need to be addressed. Future work should focus on developing strategies to preserve the conversational capabilities of fine-tuned models and reduce instances of hallucination. Overall, this method shows great potential for improving the efficiency and effectiveness of LLMs, but it requires careful consideration and further innovation to fully realize its benefits.

\begin{ack}


This work was partially supported by China Postdoctoral Science Foundation (2023M733654),  Guangdong Basic and Applied Basic Research Foundation (2023A1515110496), Shenzhen Science and Technology Innovation Program (KQTD20190929172835662).

\end{ack}


\bibliographystyle{splncs04}
\bibliography{anthology,custom}
\end{CJK*}
\end{document}